\documentclass[letterpaper, 10pt, conference]{IEEEtran}  %

\IEEEoverridecommandlockouts
\usepackage{float} 
\usepackage{amsmath,amssymb,amsfonts}
\usepackage{multirow}
\usepackage{algorithm}
\usepackage{algorithmic}
\usepackage{graphicx}
\usepackage{textcomp}
\usepackage{xcolor}
\usepackage{cite}
 \usepackage{booktabs} 
\def\BibTeX{{\rm B\kern-.05em{\sc i\kern-.025em b}\kern-.08em
    T\kern-.1667em\lower.7ex\hbox{E}\kern-.125emX}}

\begin{document}

\title{Generalizable Trajectory Prediction via Inverse Reinforcement Learning with Mamba-Graph Architecture}


\author{Wenyun Li$^{1,2}$, Wenjie Huang$^{2,3}$, Zejian Deng$^{2}$ and Chen Sun$^{2,*}$
	\thanks{*This work was not supported by any organization}
    \thanks{$^{1}$ Department of Mathematics, The University of Hong Kong (HKU)}
	\thanks{$^{2}$ Department of Data and Systems Engineering, HKU}
    \thanks{$^{3}$ Musketeers Foundation Institute of Data Science, HKU}
    \thanks{*Corresponding author: Chen Sun \tt\small c87sun@hku.hk}%
}

\maketitle

\begin{abstract}
Accurate driving behavior modeling is fundamental to safe and efficient trajectory prediction, yet remains challenging in complex traffic scenarios. This paper presents a novel Inverse Reinforcement Learning (IRL) framework that captures human-like decision-making by inferring diverse reward functions, enabling robust cross-scenario adaptability.
The learned reward function is utilized to maximize the likelihood of output by integrating Mamba blocks for efficient long-sequence dependency modeling with graph attention networks to encode spatial interactions among traffic agents.
Comprehensive evaluations on urban intersections and roundabouts demonstrate that the proposed method not only outperforms various popular approaches in terms of prediction accuracy but also achieves 2.3 times higher generalization performance to unseen scenarios compared to other baselines, {achieving adaptability in Out-of-Distribution settings that is competitive with fine-tuning.}
\end{abstract}


\section{Introduction}
Modern intelligent transportation systems (ITS) rely on accurate trajectory prediction to enhance road safety and traffic efficiency \cite{sun2023toward}. However, predicting vehicle trajectories in complex urban environments remains a significant challenge due to the intricate interplay of factors, including dynamic road geometries, dense traffic interactions, and the inherent stochasticity of human driving behavior \cite{sun2022operational}. 
{Beyond these challenges, deploying trajectory prediction in autonomous driving requires addressing the need for fast and accurate trajectory prediction in unseen, even Out-of-Distribution (OOD) \cite{Shen2021TowardsOG} scenarios. This necessitates handle the domain shift between training dataset and real-world deployment conditions.}

\par
Traditional prediction models with supervise learning paradigms aim to learn the mapping from historical trajectories to future trajectories, as well as the hidden state of the environment. Sequence modeling approaches, such as recurrent neural networks (RNNs) \cite{Zyner2018ARN} and long short-term memory (LSTM) networks \cite{Park2018SequencetoSequencePO}, have been widely employed to model temporal dependencies in trajectory data. 
With the advent of Mamba \cite{Gu2023MambaLS}, a selective State Space Model (SSM), it offers superior capability in modeling long-range dependencies while maintaining exceptional inference efficiency—a critical attribute for real-world autonomous driving applications where computational latency is non-negligible.
However, these methods often fall short in accurately capturing the intricate spatio-temporal interactions between vehicles and their environments. Graph-based encoding methods, such as Graph Neural Networks (GNNs) have emerged as a promising solution to address the interrelationships between vehicles and their surroundings \cite{Westny2023MTPGOGP}. Yet, these models often struggle to capture the nuanced, context-dependent decision-making processes of human drivers, leading to suboptimal or unsafe predictions, particularly in scenarios involving complex maneuvers or unexpected events. Thus in this paper, we combined Mamba block and GNN in our predictor module. 

\par
To further address the challenges of dataset dependency and cross-scenario generalization, \textit{Inverse Reinforcement Learning (IRL)} has shown promise in trajectory prediction complementing these supervised learning approaches.
Whereas pure IRL approaches \cite{Huang2020DrivingBM}, constrained by their reliance on environment dynamics modeling, are only viable in simplistic settings, our IRL-augmented data-driven approach not only extends applicability to complex environments, but also preserves high predictive accuracy. Diverging from conventional paradigms \cite{Geng2023MultimodalVT}—which hinge on maximum margin theory and iteratively refine policies through learner-expert performance comparisons, thus viable solely for tasks with limited policy diversity—this work embraces the maximum entropy (MaxEnt) IRL to yield policies of maximal diversity, thereby conferring robustness against the noise and uncertainty endemic to human demonstrations, while simultaneously sustaining adaptability to dynamic driving scenarios.

\begin{figure}[t]
\centerline{\includegraphics[width=.95\linewidth]{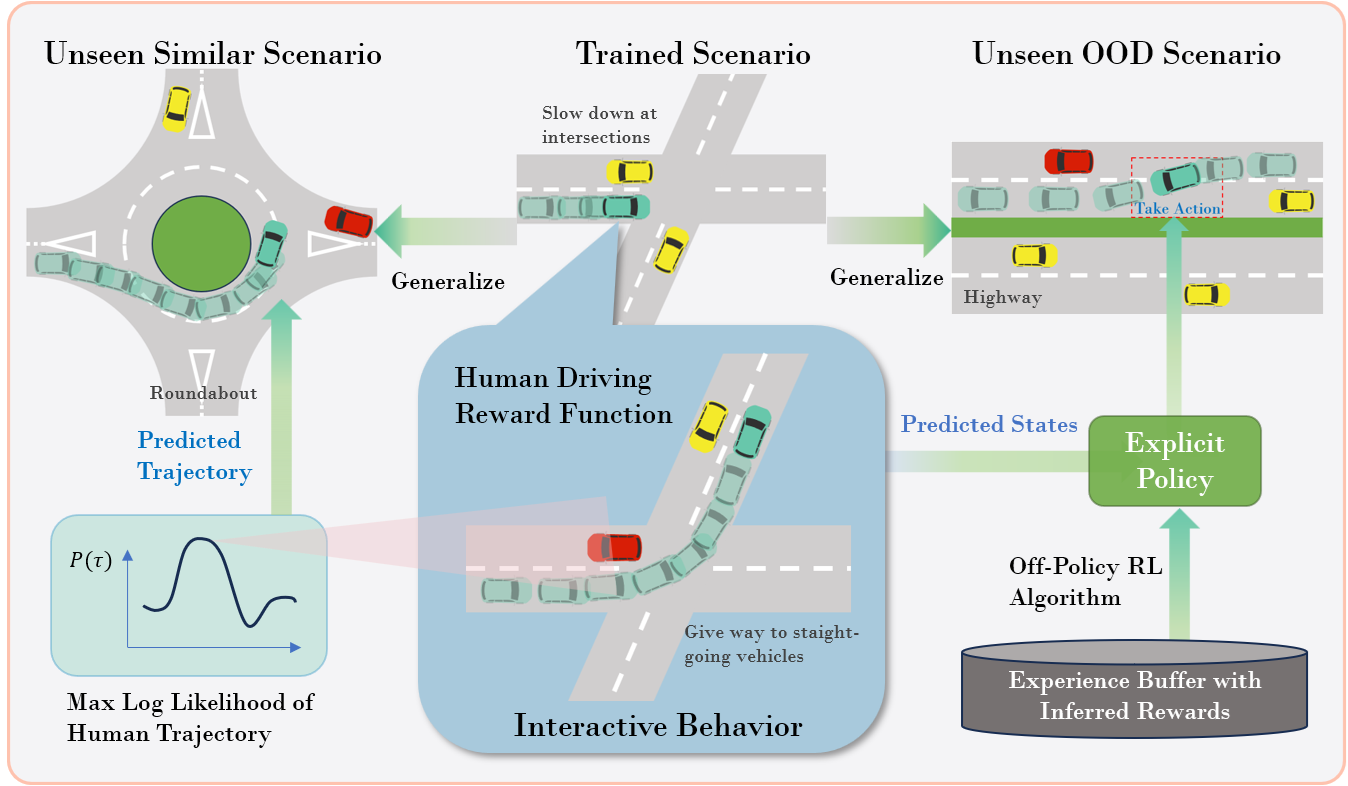}}
\caption{The illustration of Maximum Entropy Inverse Reinforcement Learning based driving behavior modeling on Cross-Scenario Adaptability.}
\label{fig.illustration}
\end{figure}

{In OOD scenarios, where the environmental data distribution completely differs from the training dataset,  most data-driven methods experience significant performance degradation. Furthermore, recent studies have highlighted the Mamba architecture's limitations in generalization, such as constrained length extrapolation \cite{benkish2024decimambaexploringlengthextrapolation} and limited domain generalization capabilities in visual tasks \cite{DGMamba}.
Despite these challenges, by leveraging the human driving reward function acquired through IRL, our approach integrates an explicit policy extension using established off-policy Reinforcement Learning (RL) algorithm. This integration substantially improves trajectory prediction performance in OOD settings,  achieving results comparable to the fine-tuned method that utilize ground truth data from new environments—all without requiring any actual ground truth data from the target scenarios (Fig.~\ref{fig.illustration}).
}

The main contributions of the paper are listed as below:
 \begin{itemize}
     \item Our method leverages the Mamba architecture to efficiently model long-range state-space dependencies in vehicle trajectories, augmented by GNN module that encodes inter-vehicle interactions through spatio-temporal message passing.
     \item We integrate a reward function module that learns human-preferential driving policies by maximizing the likelihood of demonstrated trajectories, endowing the model with generalization capabilities in analogous scenarios.
     \item The proposed algorithm outperforms state-of-the-art methods on benchmark datasets while demonstrating superior cross-scenario generalization in specialized tests, validating its practical viability for real-world autonomous driving applications.
     \item {
    We extend the generalization capability of our method by incorporating an explicit policy submodule, which achieves OOD adaptability comparable to fine-tuning approaches, all without substantial computational overhead.
     }
 \end{itemize}

The rest of paper is organized as follows: Section~\ref{section.method} presents the proposed framework, detailing its core modules. Section~\ref{section.exp} describes the experimental setup, presents validation results, and provides in-depth analysis of the findings. Section~\ref{section.conclusion} concludes the paper with a summary of key contributions and discusses promising directions for future research.

\section{Methodology}\label{section.method}
\begin{figure}[htbp]
\centerline{\includegraphics[width=0.9\linewidth]{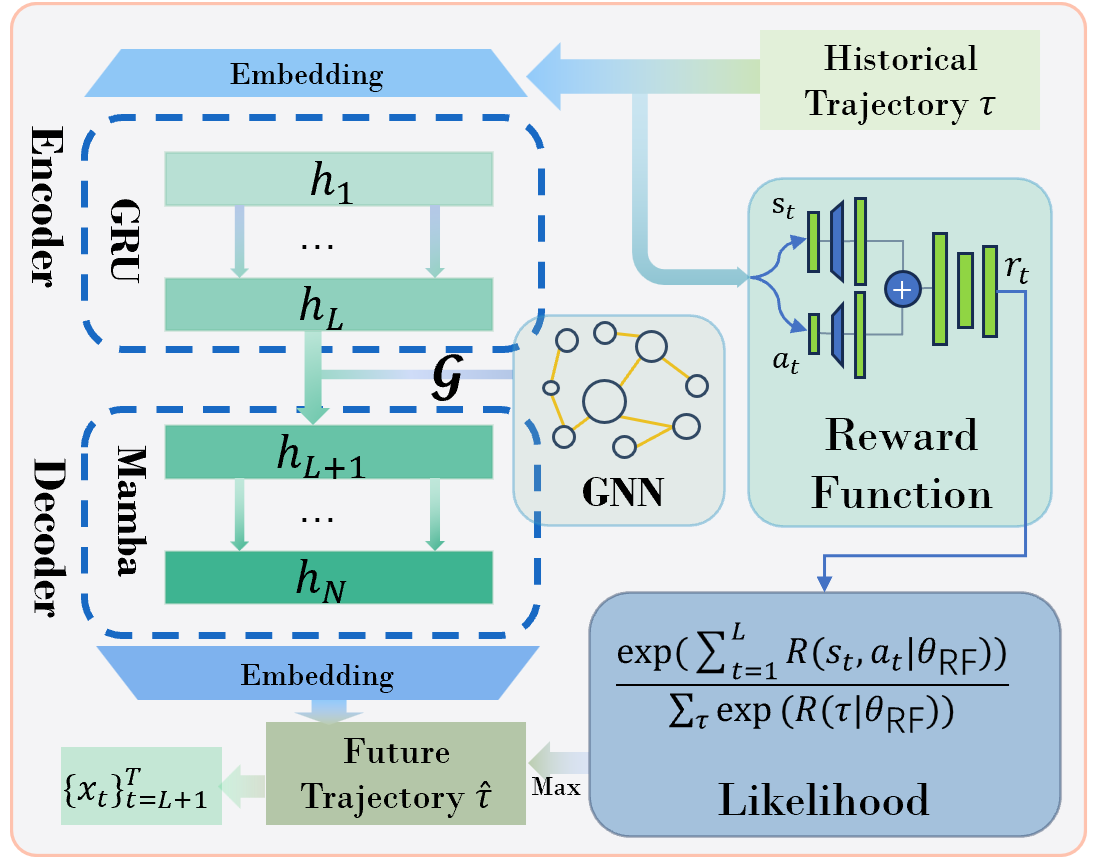}}
\caption{The overall structure of the proposed trajectory prediction framework.}
\label{fig.2}
\end{figure}

This section presents our proposed framework (Fig.~\ref{fig.2}), Environment-Aware Mamba Trajectory Predictor with MaxEnt IRL, which integrates sequence modeling with IRL to capture human driver behavior. We first formalize the trajectory prediction problem, then detail our encoder-decoder architecture leveraging Mamba blocks for efficient long-sequence modeling and Graph Attention Network (GAT) to encode spatial interactions. Finally, we introduce the MaxEnt IRL module to infer driver reward functions from heterogeneous human demonstrations, enabling more human-like and scenario-adaptive trajectory generation.

\subsection{Problem Statement}\label{Problem Statement}
We employ $\tau$ to represent the historical trajectory of the target vehicle, while $\hat{\tau}$ denotes its future trajectory. Note that the analysis in this paper assumes a discrete framework, where a trajectory is defined as a sequence of states, i.e., $\tau = [s_1, s_2, \dots, s_L]$, with $L$ indicating the length of the given trajectory and $s_t$ representing the state of the vehicle at the $t$-th time step  such as position, velocity, yaw angle, acceleration. The discussion of trajectory representation will be extended to include inter-vehicle information expressed through graph-theoretic formalism in Section~\ref{section.TPM}. 

While the RL-based prediction involves uncertain approximations of the environmental dynamics, the historical trajectory input, which, even as time series data, implicitly encompasses state-action pairs, i.e., $a_t=\Delta x_t=x_{t+1}-x_t$ where $x_t$ is the coordinate position of the target vehicle contained in $s_t$. This treatment abstracts the decision-making to a behavioral level, capturing high-level intent and social interactions rather than low-level control signals. Under high-frequency sampling, displacement corresponds to instantaneous velocity, and we implicitly assume a near-ideal underlying controller—making the mapping from intent $a_t$ to state change $s_{t+1} - s_t$ effectively deterministic. Thus, geometric displacement is justified as a valid “action,” enabling the use of an MDP formulation and IRL analysis as detailed in Section~\ref{sec.IRL}.

Our goal is to predict the future trajectory of target vehicle, specifically the coordinate position $\{\hat{x_t}\}^T_{t=L}$, given historical trajectory $\tau\in \Gamma_S$, where $T$ is the final step of the prediction.
Furthermore, we aims to implicitly infer a policy $\pi^*$ that maximizes the likelihood of generated trajectories $\hat{\tau} \sim \pi^*$,$\hat{\tau} \in \Gamma_T$, where $\Gamma_S$ and $\Gamma_T$ are the source and target domain of trajectory data. The potential discrepancy between source and target domains provides the fundamental motivation for investigating our method's cross-scenario adaptability.

\subsection{Graph-Based Mamba Predictor}\label{section.TPM}
Our prediction module adopts the widely utilized encoder-decoder architecture, wherein the embedded historical trajectory $\tau$ is processed by the encoder to yield a latent representation $h$ that encapsulates the spatio-temporal context, which subsequently serves as the input to the decoder for generating the trajectory $\hat{\tau}$. 
In the proposed architecture, the Mamba model is exclusively employed as the decoder, a deliberate design choice predicated on the observation that its formidable temporal representational capacity---while advantageous for sequential generation---would render an end-to-end fully Mamba-based encoder-decoder architecture prohibitively challenging to optimize.  

The State Space Models map an input signal $u(t) \in \mathbb{R}^D$ to the output signal $y(t) \in \mathbb{R}^D$ via the latent state $h(t) \in \mathbb{R}^N$.
For application to a discrete autoregressive input sequence of target vehicle states  $(s_0, s_1, \dots)$ from each preceding timestep, our Mamba module iteratively updates its hidden state $h_t$ at each step $t$ (where $t$ ranges from $L+1$ to $T$):
\begin{equation}\label{eq.mamba3}
    h_t = \bar{A} h_{t-1} + \bar{B} \hat{s}_{t-1},
\end{equation}
\begin{equation}\label{eq.mamba4}
    \hat{s}_t = C h_t,
\end{equation}
where $ \bar{A} , \bar{B}\text{ and }C$ are discrete trainable matrices. 
By converting the continuous SSM to the discrete one using the zero-order hold 
discretization rule, the model becomes a sequence-to-sequence mapping framework. 
Through Eq.~(\ref{eq.mamba3}) and (\ref{eq.mamba4}), the predicted sequence of states $\{\hat{s_t}\}^T_{t=L}$ is recursively computed from each hidden state $h_t$, and thus obtaining the future coordinate position $\{\hat{x_t}\}^T_{t=L}$.

To complete the description of our encoder-decoder architecture, the encoder employs a Gate Recurrent Unit (GRU) to integrate inter-vehicle information into the {top of GRU output-}latent representation $h_L$, through a GAT. Inter-vehicle relationships are modeled as graph $\mathcal{G}= (\mathcal{V}, \mathcal{E})$, where $\mathcal{V}$ denotes the node set, $\mathcal{E}$ denotes the edge set. Each node represents a traffic participant. 
For feature representation, let $h^\nu \in \mathbb{R}^{d_f}$ be the features of node $\nu \in \mathcal{V}$, $e^{\nu, \iota} \in \mathbb{R}^{d_e}$ be the features of edge $(\nu, \iota) \in \mathcal{E}$. 
This design choice is particularly crucial as the relational information between target and surrounding vehicles is typically provided as the node and edge features in the heterogeneous graph.

Within this framework, we implement an attention mechanism that computes aggregation weights across each node $\nu$'s inclusive neighborhood $\tilde{\mathcal{N}}(\nu)=\mathcal{N}(\nu) \cup \{\nu\}$, enabling the model to allocate varying importance to different neighbors. Specifically, we utilize an enhanced attention computation where the attention weights $\alpha_{\nu,\iota}$ are determined by:
\begin{equation}
    \alpha_{\nu,\iota} = \frac{\exp\left(\mathbf{a}_{\alpha}^\top \gamma\left(\mathbf{W}_{\alpha}[h^\nu \| h^\iota \| e_{\nu,\iota}]\right)\right)}{\sum_{u \in \tilde{\mathcal{N}}(\nu)} \exp\left(\mathbf{a}_{\alpha}^\top \gamma\left(\mathbf{W}_{\alpha}[h^\nu \| h^u \| e_{\nu,u}]\right)\right)},
\end{equation}
where $\mathbf{a}_{\alpha}$ and $\mathbf{W}_{\alpha}$ are learnable parameters of GATs;
$\gamma(\cdot)$ denotes the Leaky ReLU activation function (with negative slope $0.2$), providing non-linear transformation;
$[h^\nu \| h^\iota][e_{\nu,\iota}]$ indicates concatenation of features. 
{With graph considered, the output of GRU, i.e., hidden state $h^{\prime\nu}$ of the target vehicle node $\nu$, is subsequently updated through:}
\begin{equation}
    h^{\prime\nu} = b+\sum_{\iota \in \tilde{\mathcal{N}}(\nu)} \alpha_{\nu,\iota} \mathbf{W}_{\alpha} h^\iota,
\end{equation}
where $b$ is a learnable bias term. 

\subsection{Mamba-based MaxEnt IRL} 
\label{sec.IRL}
As delineated in Section~\ref{Problem Statement}, given $N$ observed trajectories $\Gamma=\{\tau_i\}^N  = \{s_1, s_2, \dots, s_L\}_N$, which can be transformed into a set of human
demonstrations $\mathcal{D}=\{(s_t,a_t)\}^{L\times N}_{t=1}$, our principal objective revolves around the estimation of the underlying reward function, the accurate reconstruction of which would enable the emulation and prediction of nuanced human driving behaviors with enhanced fidelity. 

Assume that the reward function of the driver is denoted by $R$ parametrized by $\theta_{\text{RF}}$. We employ a neural network with activation layers to fit the reward function for better performance instead of a linear structured reward function in \cite{Huang2020DrivingBM}. According to the principle of MaxEnt IRL \cite{Ziebart2008MaximumEI}, we learn the reward function by minimizing the negative log-likelihood of the human trajectories, as shown in Eq.~(\ref{eq.loss_expert}). 

\begin{equation}\label{eq.loss_expert}
    \mathcal{L}_{\text{RF}} = -\mathbb{E}_{\tau \sim \pi^*} \left[\log P(\tau|\theta_{\text{RF}})\right],
\end{equation}
where $\pi^*$ is the human strategy within the human demonstration. 
The trajectory likelihood $P(\tau)$, governed by the maximum entropy principle, is proportional to the exponential of the trajectory's cumulative reward.
To elaborate with precision, the likelihood of a trajectory $\tau$ necessitates normalization through division by the partition function $Z(\theta_{\text{RF}})$ as defined in Eq.~(\ref{eq.Z}). 
Here with a bit abuse of notation, 
the summation $\sum_{\tau}$ enumerates over the entire space of feasible human trajectories, while $R(\tau|\theta_{\text{RF}})$ encapsulates the cumulative rewards associated with a specific trajectory $\tau$ parameterized by $\theta_{\text{RF}}$. 
This normalization not only guarantees the requisite properties of a probability measure but also introduces a logarithmic regularization term during optimization, thereby mitigating the risk of overfitting to the human demonstrations. 
\begin{equation}
\begin{split}
        P(\tau|\theta_{\text{RF}}) &= \frac{1}{Z(\theta_{\text{RF}})}\exp\left( \sum_{t=1}^{L} R(s_t, a_t|\theta_{\text{RF}}) \right) \\
        &= \frac{\exp\left( \sum_{t=1}^{L} R(s_t, a_t|\theta_{\text{RF}}) \right)}
        {\sum_{\tau} \exp\left(R(\tau|\theta_{\text{RF}})\right)}.
\end{split}
\end{equation}

Directly computing the partition function $Z$ over the vast trajectory space is typically infeasible. To circumvent this, we approximate $Z$ via importance sampling using Monte Carlo estimation with $M$ samples from $\Gamma$:
\begin{equation}\label{eq.Z}
    \hat{Z}(\theta_{\text{RF}}) \approx \frac{1}{M} \sum_{t=1}^{M} \frac{e^{R(s_t,a_t|\theta_{\text{RF}})}}{\pi^*(a_t|s_t)}.
\end{equation}
When the optimal policy $\pi^*$ remains unspecified and the forward derivation of agent policy is beyond the scope of current investigation, we adopt the simplifying assumption of uniform exploration, where $\pi^*(a|s)$ is treated as constant proportional to the trajectory length.

To prevent overfitting of the reward function, a regularization term is added to constraint the norm of reward function parameter $\theta_{\text{RF}}$, yielding the final instantiation of the reward function loss:
\begin{equation}\label{eq.RF}
    \mathcal{L}_{\text{RF}} = -\mathbb{E}_{\tau \sim \pi^*} \left[\log P(\tau|\theta_{\text{RF}})\right]
    +\lambda \| \theta_{\text{RF}} \|^2.
\end{equation}

The Trajectory Prediction Module (TPM) integrates all components described in Section~\ref{section.TPM}, namely the graph-based encoder and Mamba decoder. The TPM updates its parameters by simultaneously optimizing the mean squared error of the predicted coordinates, while also maximizing the cumulative reward of the predicted trajectory in accordance Eq.~(\ref{eq.loss_expert}): 
\begin{equation}\label{eq.TPM}
\begin{aligned}
        \mathcal{L}_{\text{TPM}} = &\frac{1}{T-L} \sum_{t=L}^{T} \left\|  x_t -  \hat{x}_t \right\|^2 \\
    &- \gamma \mathbb{E}_{{\hat\tau} \sim \pi^*} \left[\log P({\hat\tau(\theta_{\text{TPM}})|\theta_{\text{RF}}})\right],
\end{aligned}
\end{equation}
where $\hat\tau(\theta_{\text{TPM}})$ is used to clarify the reliance of predicted trajectory and the parameter of the TPM.  
This integrated approach enables the graph-based Mamba predictor to generate trajectory predictions while the reward function module ensures likelihood maximization of these trajectories. The complete workflow is presented in Algorithm.~\ref{alg:1}.

\begin{algorithm}[h]
    \renewcommand{\algorithmicrequire}{\textbf{Input:}}
	\renewcommand{\algorithmicensure}{\textbf{Output:}}
\caption{Mamba predictor with MaxEnt IRL}
\label{alg:1}
\begin{algorithmic}[1]
\REQUIRE Human trajectories $\Gamma=\{\tau_i\}^N$, $\theta_{\text{TPM}}$, $\theta_{\text{RF}}$, Heterogeneous Graph $\mathcal{G} = \left( 
        \{\mathbf{X}^t\}_{t \in \mathcal{T}_v}
        \{\mathbf{E}^r, \text{Edge Index}^r\}_{r \in \mathcal{T}_e}
\right)$

\ENSURE Optimized trajectory prediction module and reward function parameters $\theta_{\text{TPM}}^*$, $\theta_{\text{RF}}^*$

\STATE \textbf{Initialize} $\theta_{\text{TPM}}$, $\theta_{\text{RF}}$ 

\FOR{epoch $\leftarrow$ 1 \TO $N$}

    \STATE Obtain $h_L$ from $\tau$ using encoder
    \STATE Encode environmental information $\mathcal{G}$ into $h_L$ via GAT
    \STATE Decode $h_L$ using Mamba to generate: \\
    \quad - The predicted trajectory $\hat{\tau}$ \\
    \quad - Future coordinates $\{\hat{x}_t\}_{t=L}^T$ 
    \STATE Compute reward values from trajectories for maximum likelihood in both $\mathcal{L}_{\text{TPM}}$ and $\mathcal{L}_{\text{RF}}$: \\
    \quad - Historical trajectory rewards $\{r_i\}_{t=1}^L=R(\tau|\theta_{RF})$ \\
    \quad - Predicted trajectory rewards $\{r_i\}_{t=L}^T=R(\hat{\tau}|\theta_{RF})$
    \STATE \text{Loss calculation by Eq.~(\ref{eq.RF}) and (\ref{eq.TPM})}:
       \begin{align*}
       \mathcal{L}_{\text{TPM}} \leftarrow & \frac{1}{T-L} \sum_{t=L}^{T} \left\|  x_t -  \hat{x}_t \right\|^2 \\
       & -\gamma \mathbb{E}_{{\hat\tau} \sim \pi^*} \left[\log P({\hat\tau(\theta_{\text{TPM}})|\theta_{\text{RF}}})\right],
       \end{align*}
     \[\mathcal{L}_{\text{RF}} \leftarrow     -\mathbb{E}_{\tau \sim \pi^*} \left[\log P(\tau|\theta_{\text{RF}})\right]
    +\lambda \| \theta_{\text{RF}} \|^2.\]
    \STATE $\theta_{\text{TPM}}^{*} \leftarrow \theta_{\text{TPM}}$
    \STATE $\theta_{\text{RF}}^{*} \leftarrow \theta_{\text{RF}}$
\ENDFOR
\end{algorithmic}
\end{algorithm}

{
\subsection{Out-of-Distribution Generalization For Mamba Predictor}

In OOD scenarios, directly predicting accurate future vehicle trajectory coordinates becomes particularly challenging. To achieve broader scenario coverage, we introduce an explicit policy extension, by employing the learned driver policy to produce sequential decisions over future states to generate predicted trajectories. Specifically, the driver agent takes the predicted future states as input, outputs predicted actions $a = \pi_{\theta_{\text{agent}}}(s)$ which align with human driving strategies, and subsequently produces predicted future vehicle coordinates. A detailed description of this process is provided in Algorithm.~\ref{alg:explicit policy}.
Although the cumulative offset operation in the diagram is a for-loop, parallel computation is implemented in trajectory level and batch level, significantly reducing processing time.
Parameter $\theta_{\text{agent}}$ is optimized using the off-policy Actor-Critic algorithm \textit{Twin Delayed Deep Deterministic Policy Gradient} (TD3) \cite{Fujimoto2018AddressingFA}.

To support this learning process, we construct an experience replay buffer with reward value labeled for each state-action pairs in vehicle trajectories using the reward function derived earlier. We further introduce a penalty term in the reward proportionally to their prediction errors (e.g., negative Mean Squared Error). Concretely, for each state-action pair with state $s_t$ and action $a_t$, we complete the transition with reward:
\begin{equation}
    r=R(s_t, a_t|\theta_{\text{RF}})-\left\|a_t-\pi_{\theta_{\text{agent}}}(s_t)\right\|^2
\end{equation}
}

\begin{algorithm}[h]
\caption{Predict Trajectories in OOD Scenario with Explicit Policy}
\label{alg:explicit policy}
    \renewcommand{\algorithmicrequire}{\textbf{Input:}}
	\renewcommand{\algorithmicensure}{\textbf{Output:}}
\begin{algorithmic}[1]
 \REQUIRE Historical OOD trajectory $\tau = [s_1, s_2, \dots, s_L]$ (each $s_t$ contains coordinate $x_t$),
Optimized $\theta_{\text{agent}}^{*}$,
$\theta_{\text{TPM}}^{*}$
\ENSURE Future coordinates $\{\hat{x}_t\}_{t=L+1}^{L+T}$

\STATE Predict future states: $s_{\text{future}} = [s_{L+1}, s_{L+2}, \dots, s_{L+T}] \gets \mathcal{P}_{\theta_{\text{TPM}}^{*}}(\tau)$
\STATE Initialize action sequence $A \gets \emptyset$
\FOR{each predicted state $s_t \in s_{\text{future}}$}
    \STATE Generate action: $a_t \gets \pi_{\theta_{\text{agent}}^{*}}(s_t)$
    \STATE Append to action sequence: $A \gets A \cup \{a_t\}$
\ENDFOR
\STATE Initialize future coordinates $\{\hat{x}_t\}_{t=L+1}^{L+T} \gets \emptyset$ and cumulative offset $\delta \gets 0$
\STATE Let $x_L$ be the coordinate from the last historical state $s_L$
\FOR{each action $a_t \in A$ with index $t$ from $L+1$ to $L+T$}
    \STATE Update cumulative offset: $\delta \gets \delta + a_t$
    \STATE Compute future coordinate: $\hat{x}_t \gets x_L + \delta$
    \STATE Add to output: $\{\hat{x}_t\}_{t=L+1}^{L+T} \gets \{\hat{x}_t\}_{t=L+1}^{L+T} \cup \{\hat{x}_t\}$
\ENDFOR
\RETURN $\{\hat{x}_t\}_{t=L+1}^{L+T}$
\end{algorithmic}
\end{algorithm}

\section{Evaluation and Results}\label{section.exp}
The efficacy of the proposed framework has been 
validated through an extensive suite of experiments, encompassing diverse driving scenarios and performance metrics. To ensure a holistic assessment, we first conduct a systematic comparison against state-of-the-art baselines across benchmark datasets. Subsequently, we present detailed ablation studies and cross-scenario test to dissect the contributions of individual components, thereby quantifying their impact on overall performance.

\begin{figure}[t]

\centerline{\includegraphics[width=0.5\linewidth]{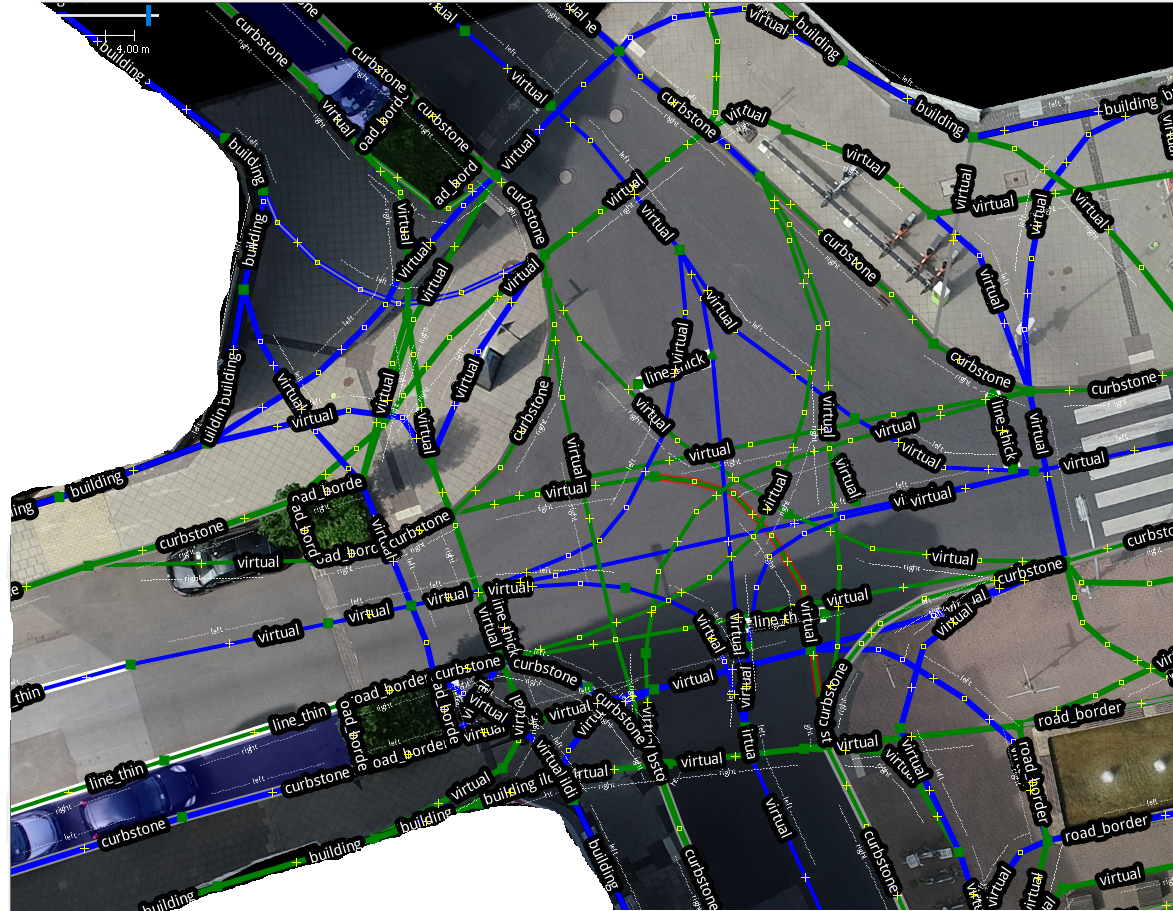}}
\caption{Environment and trajectory notation of urban intersection scenario at Heckstrasse \cite{Bock2019TheID}.}
\label{fig.inD_visual}
\end{figure}

\vspace{0.5cm}\noindent\textbf{Datesets.} Our experimental evaluation leverages {four large-scale trajectory datasets---the \textit{rounD}~\cite{Krajewski2020TheRD}, \textit{inD}~\cite{Bock2019TheID}, \textit{highD}~\cite{Krajewski2018TheHD} and \textit{exiD}~\cite{exiDdataset} datasets---which comprise vehicular movement patterns captured at 25Hz across diverse German roundabout, intersectio (Fig.~\ref{fig.inD_visual}), highway entries and exits scenarios.} 
These datasets, having undergone preprocessing procedures outlined in~\cite{Westny2023MTPGOGP}, are subsequently downsampled by a factor of 5 to achieve a temporal resolution of 0.2s. 
{The observation window and prediction horizon are set to 3s ($L$=15) and 5s ($N$=40) for the \textit{inD} and \textit{rounD} datasets, and to 2s and 5s for the \textit{highD} and \textit{exiD} datasets, respectively.
The data is split into a stratified 80-10-10 partition for training, validation, and testing.

Although IRL is potentially sensitive to the quality of demonstrations, we made no attempts to filter or curate the data to avoid inconsistent or unsafe behaviors.
Note that in all cross-scenario testing described herein, models are trained only on source domain data and evaluated solely on target domain data.
Highway scenarios (including entries and exits) serve as OOD test environments, characterized by fundamentally distinct probability distributions of lane-changing, turning, acceleration, and braking behaviors compared to the source domain of urban intersections and roundabouts, as substantiated by the baseline evaluation results in OOD test (Tab.~\ref{tab.ood}).

}

\vspace{0.5cm}\noindent\textbf{Metrics} The algorithm's predictive performance was evaluated using five key metrics: Average Displacement Error (ADE), Final Displacement Error (FDE), Average Path Displacement Error (APDE), Miss Rate (MR), and Crash Rate (CR). Their definitions follow \cite{Westny2023MTPGOGP}.

\begin{table}[ht]
\caption{RounD Dataset Vehicular Trajectory Prediction Metrics}
\label{tab.rounD}
\centering
\begin{tabular}{lcccc}
\toprule
\textbf{Model} & \textbf{ADE} & \textbf{FDE} & \textbf{MR} & \textbf{APDE} \\
\midrule
\multicolumn{5}{l}{\textit{RounD}} \\
\midrule
CA & 4.83 & 16.2 & 0.95 & 3.9 \\
CV & 6.49 & 17.1 & 0.94 & 4.34 \\
Seq2Seq & 1.46 & 3.66 & 0.59 & 0.82 \\
S-LSTM \cite{Alahi2016SocialLH} & 1.2 & 3.47 & 0.56 & 0.74 \\
CS-LSTM \cite{Deo2018ConvolutionalSP} & 1.19 & 3.57 & 0.6 & 0.69 \\
GNN-RNN \cite{Mo2021GraphAR} & 1.29 & 3.5 & 0.59 & 0.77 \\
mmTransformer \cite{Liu2021MultimodalMP} & 1.29 & 3.5 & 0.59 & 0.77 \\
Trajectron++ \cite{Salzmann2020TrajectronDT} & 1.09 & 3.53 & 0.54 & 0.59 \\
MTP-GO \cite{Westny2023MTPGOGP} & 0.96 & 2.95 & 0.46 & 0.59 \\
Mixed Mamba \cite{Zhang2024EnhancedPO} & 0.93 & 3.03 & \textbf{0.27} & 0.57 \\
Ours & \textbf{0.91} & \textbf{2.78} & 0.43 & \textbf{0.57} \\
\bottomrule
\end{tabular}
\end{table}

\begin{table}[ht]
\centering
\caption{Vehicular Trajectory Prediction Metrics on Cross-Scenario Adaptability}
\label{tab.trans}
\begin{tabular}{lccccc}
\toprule
\textbf{Model} & \textbf{ADE} & \textbf{FDE} & \textbf{MR} & \textbf{APDE} & \textbf{CR} \\
\midrule
\multicolumn{6}{l}{\textit{inD }(Seen Scenario)} \\
\midrule
GNN-RNN \cite{Mo2021GraphAR} & 0.99 & 2.84 & 0.39 & 0.55 & 0.0023 \\
mmTransformer \cite{Liu2021MultimodalMP} & 0.95 & 2.48 & 0.38 & 0.59 & 0.0024 \\
AppL \cite{Choi2019RegularizingNN} & 0.78 & 2.08 & 0.32 & 0.48 & 0.0023 \\
Ours & \textbf{0.70} & \textbf{1.95} & \textbf{0.30} & \textbf{0.42} & \textbf{0.0023} \\
\midrule
\multicolumn{6}{l}{\textit{rounD }(Unseen Scenario)} \\
\midrule
GNN-RNN \cite{Mo2021GraphAR} & 5.89 & 13.98 & 0.95 & 3.01 & 0.0038 \\
mmTransformer \cite{Liu2021MultimodalMP} & 5.36 & 11.79 & \textbf{0.92} & 2.67 & 0.0020 \\
AppL \cite{Choi2019RegularizingNN} & 5.46 & 12.19 & 0.95 & 3.97 & 0.12 \\
Ours & \textbf{4.72} & \textbf{10.27} & 0.94 & \textbf{2.42} & \textbf{0.00069} \\
\bottomrule
\end{tabular}
\end{table}

\begin{figure}[htbp]
\centerline{\includegraphics[width=0.6\linewidth]{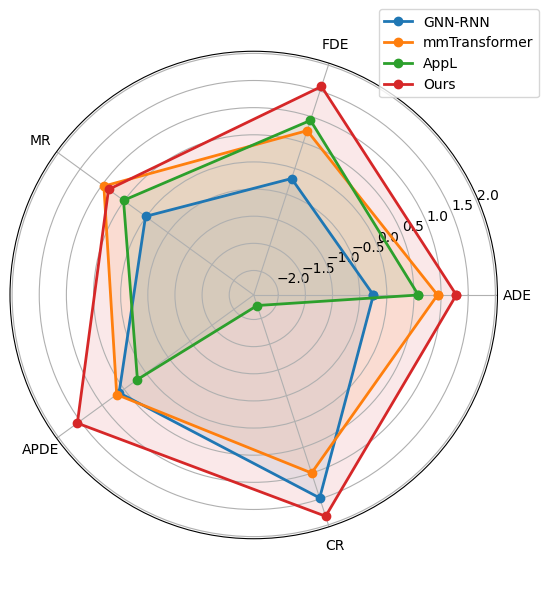}}
\caption{Cross-Scenario Adaptability (CSA) scores by metric visualized in radar chart.}
\label{fig.CSA}
\end{figure}

\begin{table*}[ht]
\centering
\caption{Ablation Experiment Results}
\label{tab.ablation}
\begin{tabular}{lccccccccc}
\toprule
\textbf{Dataset} & \textbf{Index} & \textbf{Mamba} & \textbf{MaxEntIRL} & \textbf{GNN} & \textbf{ADE} & \textbf{FDE} & \textbf{MR} & \textbf{APDE} & \textbf{CR} \\
\midrule
\multirow{4}{*}{\textit{inD}} 
& $\text{H}_1$ & \checkmark & \checkmark & \checkmark & \textbf{0.71} & \textbf{1.96} & \textbf{0.31} & \textbf{0.43} & \textbf{0.0023} \\
& $\text{H}_2$ & \checkmark & \checkmark & & 0.78 & 2.08 & 0.32 & 0.47 & 0.0031 \\
& $\text{H}_3$ & \checkmark & & \checkmark & 0.72 & 1.96 & 0.30 & 0.44 & 0.0023 \\
& $\text{H}_4$ & \checkmark & & & 0.83 & 2.13 & 0.44 & 0.49 & 0.0035 \\
\midrule
\multirow{4}{*}{\textit{rounD}}
& $\text{H}_1$ & \checkmark & \checkmark & \checkmark & \textbf{0.91} & \textbf{2.78} & \textbf{0.43} & \textbf{0.57} & \textbf{0.00067} \\
& $\text{H}_2$ & \checkmark & \checkmark & & 1.05 & 2.98 & 0.50 & 0.66 & 0.0009 \\
& $\text{H}_3$ & \checkmark & & \checkmark & 1.01 & 2.84 & 0.45 & 0.63 & 0.00074 \\
& $\text{H}_4$ & \checkmark & & & 1.06 & 3.01 & 0.51 & 0.66 & 0.00087 \\
\midrule
\multirow{4}{*}{$\textit{rounD}^*$}
& $\text{H}_1$ & \checkmark & \checkmark & \checkmark & \textbf{4.72} & \textbf{10.27} & \textbf{0.94} & \textbf{2.42} & 0.00069 \\
& $\text{H}_2$ & \checkmark & \checkmark & & 4.98 & 10.65 & 0.94 & 3.67 & 0.00069 \\
& $\text{H}_3$ & \checkmark & & \checkmark & 8.64 & 17.94 & 0.99 & 5.37 & 0.0007 \\
& $\text{H}_4$ & \checkmark & & & 7.03 & 14.74 & 0.99 & 5.01 & \textbf{0.00065} \\
\bottomrule
\end{tabular}
\end{table*}

\subsection{
Results Analysis on \textit{rounD}}

The quantitative results are presented in Tab.~\ref{tab.rounD}. 
In contrast to Mixed Mamba \cite{Zhang2024EnhancedPO}—a contemporaneous approach that similarly incorporates the Mamba module—the cited method purportedly achieves superior accuracy on the \textit{highD} dataset compared to the \textit{rounD} dataset. 
Our method achieves state-of-the-art performance across key metrics—including ADE, FDE and APDE, e.g., FDE of 2.78, representing an 8\% reduction compared to the Mixed Mamba, thereby demonstrating its robustness in handling high-variability driving behaviors. 
The discrepancy of Mixed Mamba performance can be attributed to the simplicity of highway trajectory prediction, where vehicular motion patterns are predominantly linear. Furthermore, the \textit{highD} dataset's extensive scale likely augments the Mamba module's performance.
However, the aforementioned work fails to extend Mamba's advantages to roundabout trajectory prediction, a domain that presents greater challenges due to the non-linear and dynamic nature of vehicular interactions. 
The result shows our proposed method addresses these limitations with a maximum cumulative reward objective. This integration enables our framework to excel in complex urban roundabout scenarios, such as these in Aachen and Alsdorf.

\subsection{The Effectiveness of MaxEnt IRL}
To evaluate the generalization capability and Cross-Scenario Adaptability (CSA) of our proposed methodology, we conducted experiments involving training on the \textit{inD} dataset while validating performance across both \textit{inD} and \textit{rounD} datasets, with quantitative results tabulated in Tab.~\ref{tab.trans}. This experimental paradigm was deliberately designed to probe the model's transferability between ostensibly similar yet distinct driving environments - while both datasets exhibit comparable microscopic behaviors such as car-following, lane-changing, and turning maneuvers within confined spatial contexts, they diverge in road geometries and traffic dynamics between intersections and roundabouts.

We calculate the CSA scores of different methods across various metrics for quantitative comparative analysis, which is formally defined as:
\begin{equation}
CSA = \underbrace{  \alpha \cdot M_{\text{known}}}_{\text{Known Scenario}} + \underbrace{  M_{\text{unknown}}}_{\text{Unknown Scenario}} - \underbrace{ \beta \cdot \mathcal{D},}_{\text{Degradation Penalty}}
\label{eq:csa}
\end{equation}
where:
\begin{align*}
M_{\text{known}} &= \frac{1}{n}\sum_{i=1}^n \left(1 - \frac{M_i^{\text{known}} - M_{\min}^{\text{known}}}{M_{\max}^{\text{known}} - M_{\min}^{\text{known}}}\right), \\
M_{\text{unknown}} &= \frac{1}{n}\sum_{i=1}^n \left(1 - \frac{M_i^{\text{unknown}} - M_{\min}^{\text{unknown}}}{M_{\max}^{\text{unknown}} - M_{\min}^{\text{unknown}}}\right), \\
\mathcal{D} &= \frac{1}{n}\sum_{i=1}^n \frac{M_i^{\text{unknown}} - M_i^{\text{known}}}{|M_i^{\text{known}}|},
\end{align*}
with $M_i$ being the $i$-th evaluation metric (e.g., ADE, FDE). The terms with ``$\max$'' and ``$\min$'' subscripts 
denote the maximum and minimum values of certain scenario respectively across all evaluated methods, respectively, while $\alpha,\beta$ represents the weighting coefficient. Intuitively, a higher CSA value indicates superior cross-scenario adaptability. 

The qualitative results are shown in Fig.~\ref{fig.CSA}.
Notably, although the CSA values for MR are comparatively lower across all methods---reflecting the inherent challenge of behavioral prediction in unseen scenarios---our approach still demonstrates state-of-the-art performance within this category. More significantly, our method outperforms all baselines in the remaining metrics, achieving 2.3 times the CSA scores (in terms of APDE) of the second-best method.

By virtue of employing MaxEnt IRL, which endows the learned policies with enhanced diversity, our method demonstrates superior performance in the \textit{inD} scenario (Tab.~\ref{tab.trans}). Specifically, the proposed approach achieves significantly lower ADE and FDE compared to the apprenticeship learning method (AppL \cite{Choi2019RegularizingNN}) that adheres to the \textit{maximum margin theory} framework. 
This empirical result substantiates our framework's enhanced capacity for holistically approximating the underlying distribution of human driving trajectories, demonstrating more faithful reproduction of naturalistic behavioral patterns compared to AppL approaches.

\begin{table}[htbp]
\centering
\caption{Computational Efficiency Comparison of Different Models}
\label{tab:model_comparison}
\begin{tabular}{lrr}
\toprule
 & \textbf{Time/Param} & \textbf{Memory/Param} \\
\textbf{Model} & \textbf{(s/1000 params)} & \textbf{(GB/1000 params)} \\
\midrule
Mamba & \textbf{0.0184} & \textbf{0.0130} \\
Transformer & 0.2255 & 0.1452 \\
GNN & 0.0737 & 0.2664 \\
\bottomrule
\end{tabular}
\end{table}
\vspace{-0.5cm}
\begin{flushleft}
\small\textit{Note: The Mamba model demonstrates the highest computational efficiency among the three models in terms of training memory usage relative to model size and inference time relative to model size. To avoid differences in inference time caused by varying pipelines, we only record the pure model inference time. Batch size is set to 128.}
\end{flushleft}

\textbf{Ablation Study.} The ablation experiments for the proposed method were conducted on the \textit{inD} and \textit{rounD} datasets, with the results summarized in Tab.~\ref{tab.ablation}. Notably, $\textit{rounD}^*$ denotes the cross-scenario evaluation.
{The Mamba module serves as the backbone component of the predictor and was not included in the ablation study. The claim of Mamba's exceptional inference efficiency is supported by the results in Tab.~\ref{tab:model_comparison}.}
Further analysis on the tabulated data reveals that both the MaxEnt IRL module and the GATs module contribute to trajectory prediction accuracy, with the latter exhibiting even more pronounced enhancement. This phenomenon can be attributed to the fact that the graph-based architecture, introduces additional heterogeneous graph data encompassing information that the IRL method doesn't contain. 
However, the IRL component does play a pivotal role in bolstering the model’s generalizability: in the $\textit{rounD}^*$ column in Tab.~\ref{tab.ablation}, the absence of the IRL module precipitates an 83\% deterioration in ADE and a more than twofold decline in FDE.

{
\subsection{Out-of-Distribution Scenarios Test}
\begin{table}[htbp]
\centering
\caption{OOD Scenarios Test Results}
\label{tab.ood}
\begin{tabular}{clccc}
\toprule
\textbf{Dataset} & \textbf{Method} & \textbf{ADE} & \textbf{APDE} & \textbf{FDE} \\
\midrule
\multirow{3}{*}{\textit{highD}} 
& Baseline & 638.01 & 580.1 & 678.35 \\
& + Policy & 290.81 & 228.8 & 377.41 \\
& LoRA & \textbf{233.68} & \textbf{177.64} & \textbf{259.68} \\
\midrule
\multirow{3}{*}{\textit{exiD}}
& Baseline & 243.95 & 198.59 & 285.2 \\
& + Policy & \textbf{89.89} & \textbf{43.41} & \textbf{112.89} \\
& LoRA & 99.04 & 52.69 & 118.94 \\
\bottomrule
\end{tabular}
\end{table}
The OOD adaptability experiment on the \textit{highD} and \textit{exiD} dataset validates our method's capability to generalize in unfamiliar OOD scenarios. Based on our graph based Mamba predictor, we conduct two sets of experiments:  a baseline model augmented with the explicit policy extension (Fig.~\ref{fig.TD3_metric_curve}), and baseline model finetuned accordingly with \textit{highD}/\textit{exiD} ground truth data using Low-Rank Adaptation (LoRA) \cite{hu2022lora}—a parameter-efficient fine-tuning (PEFT) method (Fig.~\ref{fig.finetune_curve}).
In the highway scenario, where kinematic models differ substantially from those in urban intersections and roundabouts, the policy extension ('+Policy' in Tab.~\ref{tab.ood}) improves the baseline performance but slightly underperforms compared to the LoRA-based fine-tuning. In contrast, on the \textit{exiD} dataset—also a highway environment but with a higher frequency of merging and exiting behaviors—the policy extension demonstrates marginally better adaptability than LoRA.

\begin{figure}[htbp]
\centerline{\includegraphics[width=0.7\linewidth]{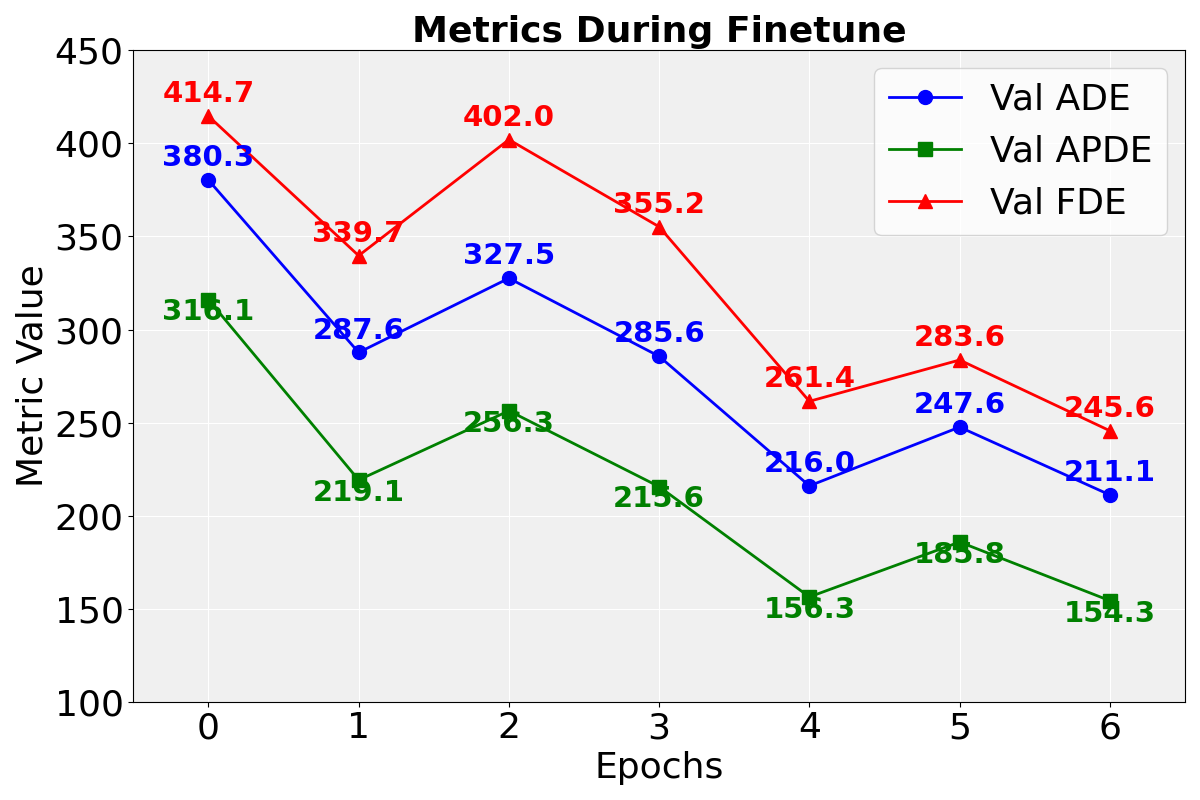}}
\caption{Metrics on validation set obtained from fine-tuning on 100 random instances of the \textit{highD} dataset, exhibiting convergence within 6 epochs. Following experimental results, the input projection, time parameter projection, and output projection modules in the Mamba block are selected as the optimal targets for LoRA fine-tuning with a parameter compression rate of 3.91\%.}
\label{fig.finetune_curve}
\end{figure}

\begin{figure}[htbp]

\centerline{\includegraphics[width=0.7\linewidth]{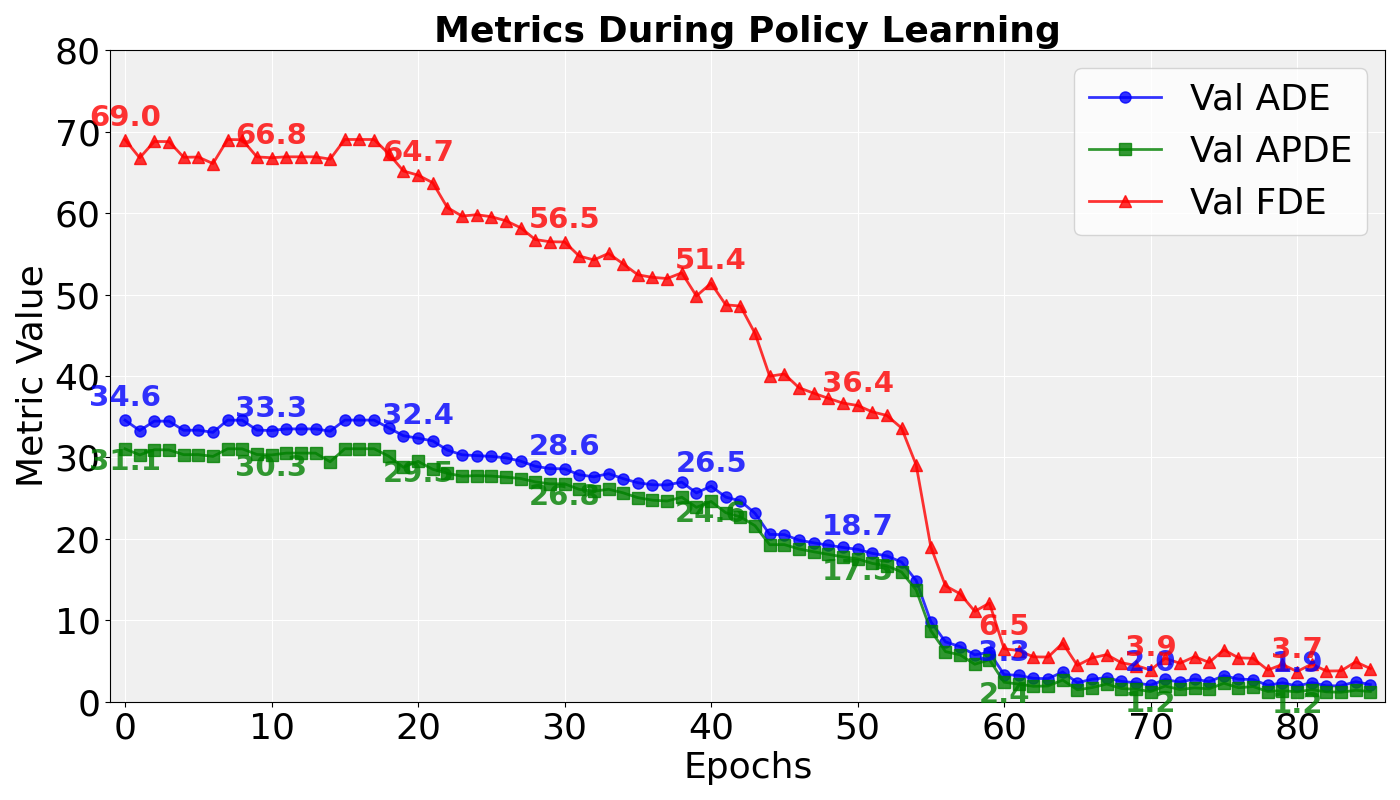}}
\caption{Validation metrics obtained during policy learning using the TD3 algorithm, showing convergence within 80 epochs. The policy delay for TD3 is set to 3. The policy is trained once in the source domain and can be directly applied to any target domain (e.g., \textit{highD} or \textit{exiD})}
\label{fig.TD3_metric_curve}
\end{figure}

The cross-scenario adaptability of our method in analogous and OOD scenario is further accentuated by the trade-off in Fig.~\ref{fig.tradeoff}: increased model complexity (larger parameters) improves in-domain accuracy but often leads to degradation of cross-scenario generalization. This result underscores the superiority of our approach for leveraging the degradation effect of introducing mamba architecture (with more parameters) and still preserves cross-scenario adaptability.

\begin{figure}[htbp]

\centerline{\includegraphics[width=0.7\linewidth]{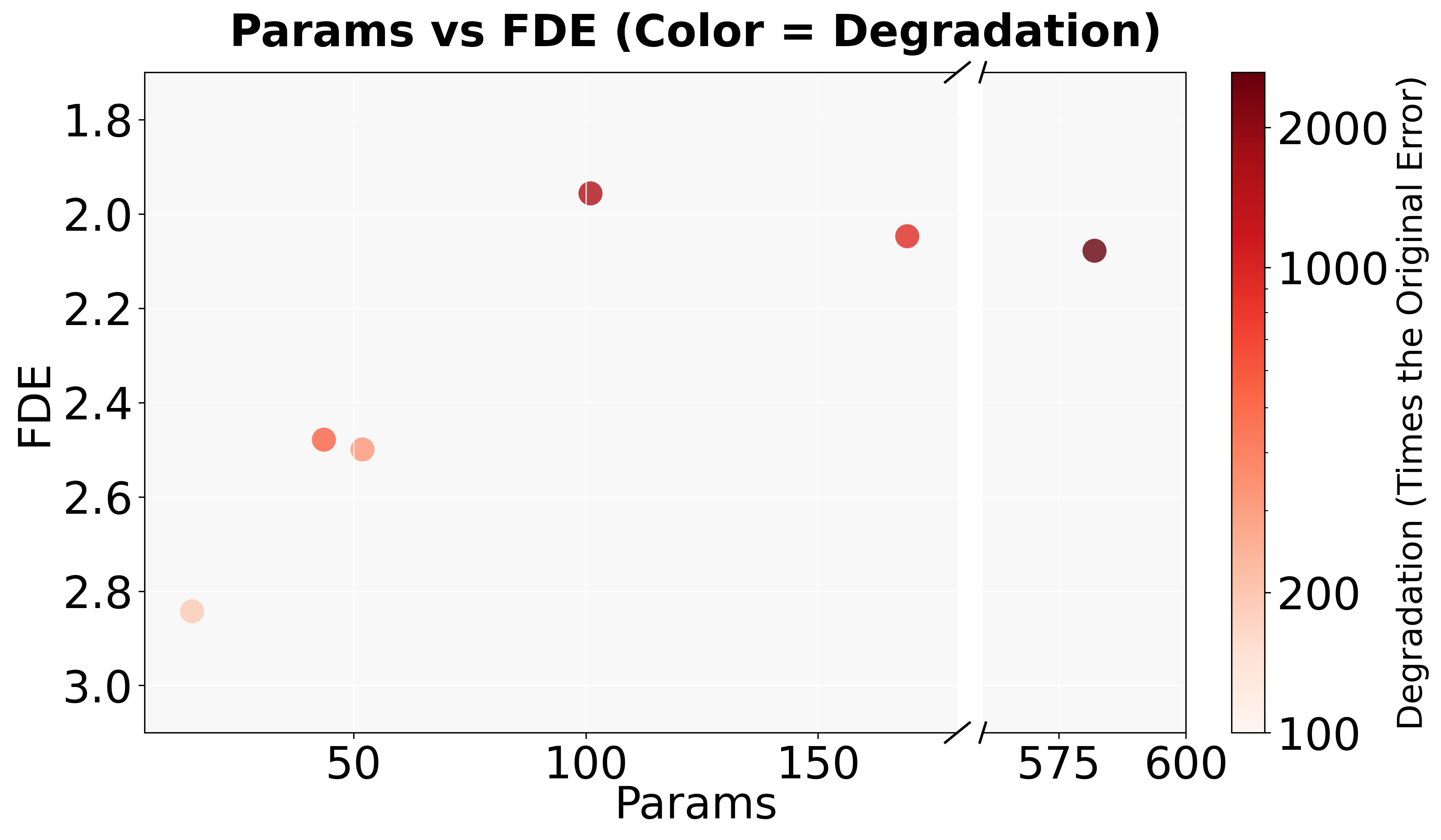}}
\caption{An illustration of the accuracy-parameter-degradation trade-off in cross-scenario adaptation (urban intersection to highway). FDE (vertical axis) is plotted against model parameter size (horizontal axis, in thousands), with the color intensity of each point representing its degree of performance degradation - darker colors represent more severe degradation.}
\label{fig.tradeoff}
\end{figure}

}

\section{Conclusion}\label{section.conclusion}
We proposed the environment-aware Mamba trajectory predictor with MaxEnt IRL, a novel framework that integrates Mamba-based sequence modeling with IRL to capture human driver behavior. We developed an encoder-decoder architecture combining Mamba blocks for efficient long-sequence dependency modeling and GATs for encoding spatial interactions among traffic agents. The MaxEnt IRL module {and policy extension} further enhanced the framework’s adaptability by inferring diverse driver reward functions from heterogeneous demonstrations, enabling human-like and scenario-aware trajectory generation.
Experiments on urban intersections, roundabouts and highways demonstrated that our method achieves state-of-the-art prediction accuracy while exhibiting strong generalization to unseen scenarios, owing to its ability to learn transferable reward structures from human data.

{Future research may explore leveraging agent policies to stabilize training through techniques like importance sampling for approximating the partition function $Z$ with $\pi^*$, and better integrate long- and short-horizon predictions to improve generalization, thereby broadening the practical applicability in real-world autonomous driving systems.}

\bibliographystyle{IEEEtran}
\bibliography{references}
\end{document}